\documentclass{llncs}

\usepackage{makeidx}
\usepackage{cite}
\usepackage{hyperref}
\usepackage[utf8]{inputenc}
\usepackage{booktabs}
\usepackage[misc]{ifsym}
\usepackage{tikz}
\usepackage{indentfirst}
\usetikzlibrary{arrows,shapes,positioning,shadows,trees,calc,decorations.markings}
\usepackage{pifont}
\newcommand{\cmark}{\ding{51}}%
\newcommand{\xmark}{\ding{53}}%
\usepackage{framed,multirow}

\begin{document}

\frontmatter
\mainmatter
\title{Learning and Reasoning with the Graph Structure Representation in Robotic Surgery}

\titlerunning{Graph Structure Representation in Robotic Surgery}
\author{Mobarakol Islam\inst{1,2} \and Lalithkumar Seenivasan\inst{1} \and Lim Chwee Ming\inst{3} \and Hongliang Ren\inst{1}}

\authorrunning{Islam et al.}
\tocauthor{Example Author}

\institute{Dept. of Biomedical Engineering, National University of Singapore, Singapore
\and
Biomedical Image Analysis Group, Imperial College London, UK
\and
Dept. of Otolaryngology, Singapore General Hospital, Singapore\\
\email{mobarakol@u.nus.edu, lalithkumar\_s@u.nus.edu, lim.chwee.ming@singhealth.com.sg, ren@nus.edu.sg/hlren@ieee.org}
\thanks{Correspondence to Prof. Ren ren@nus.edu.sg/hlren@ieee.org. This work supported by NMRC Bedside \& Bench under grant R-397-000-245-511.}
}
\maketitle
\begin{abstract}
Learning to infer graph representations and performing spatial reasoning in a complex surgical environment can play a vital role in surgical scene understanding in robotic surgery. For this purpose, we develop an approach to generate the scene graph and predict surgical interactions between instruments and surgical region of interest (ROI) during robot-assisted surgery. We design an attention link function and integrate with a graph parsing network to recognize the surgical interactions. To embed each node with corresponding neighbouring node features, we further incorporate SageConv into the network. The scene graph generation and active edge classification mostly depend on the embedding or feature extraction of node and edge features from complex image representation. Here, we empirically demonstrate the feature extraction methods by employing label smoothing weighted loss. Smoothing the hard label can avoid the over-confident prediction of the model and enhances the feature representation learned by the penultimate layer. To obtain the graph scene label, we annotate the bounding box and the instrument-ROI interactions on the robotic scene segmentation challenge 2018 dataset with an experienced clinical expert in robotic surgery and employ it to evaluate our propositions.

\end{abstract}

\section{Introduction}

In modern surgical practices, minimally invasive surgeries (MIS) has become a norm as it greatly reduces the trauma of open surgery in patients and reduces their recovery period. The accuracy and reliability of the MIS, and dexterity of a surgeon improve with the introduction of robot-assisted MIS (RMIS)\cite{okamura2010haptics}. Although RMIS holds great potential in the medical domain, its applications and capabilities are still limited by the lack of haptic feedback, reduced field of view, and the deficit of evaluation matrix to measure a surgeon's performance\cite{okamura2010haptics, laina2017concurrent, allan2012toward, lee2020ultrasound}. In recent times, attempts have been made to address these limitations by introducing new techniques inspired by advancement in the computer vision field. To address the shortcomings and assist the surgeons better, surgical tool tracking \cite{allan2012toward, islam2020ap} and instrument segmentation \cite{laina2017concurrent, pakhomov2019deep} have been introduced to allow the overlay of the medical image on top of the surgical scene. While considerable progress has been made in detecting the tools and semantic segmentation, few key problems still exist. Firstly, despite identifying the instruments, the system is unable to decide when to stimulate haptic sense. Secondly, the system cannot determine when to overlay the medical image and when not to during the surgery to ensure that it does not obstruct the surgeon's view of the surgical scene. Finally, the lack of standardized evaluation matrix for surgeons and the inability of the system to perform spatial understanding are the key reasons for these limitations.  

Spatial understanding, termed as deep reasoning in the artificial intelligence (AI) domain, enables the system to infer the implicit relationship between instruments and tissues in the surgical scene. Incorporating reasoning capability in robotic surgery can empower the machine to understand the surgical scene and allows the robot to execute informed decisions such as stimulating a haptic sense or overlay medical images over to assist surgeons based on interactions between the tissue and tool. Furthermore, this would allow the system to identify the tasks performed by surgeons, which can then be used for evaluating their performance. However, the detection of tool-tissue interaction poses one major problem. Most of the advancements made in the AI domains, such as convolutional neural networks (CNNs), addresses issues in the euclidean domain\cite{zhou2018graph}. However, detecting interactions between tissue and tools can fall under non-euclidean space where defective tissues can be related to multiple instruments, and each instrument could have multiple interaction types. For the machine learning problems in the non-euclidean domain, graph neural network (GNN) has been proposed, which allows the network to perform spatial reasoning based on features extracted from the connected nodes\cite{zhou2018graph,qi2018learning}. Inspired by graph parsing neural network (GPNN)\cite{qi2018learning}, here, we present an enhanced GNN model that infers on the tool-tissue interaction graph structure to predicts the presence and type of interaction between the defective tissue and the surgical tools.
Our key contributions in this work are as follows:
\begin{itemize}
    \item[--] Label smoothened features: Improve model performance by label smoothened node and edge features.
    \item[--] Incorporate SageConv\cite{hamilton2017inductive}: An inductive framework that performs node embedding based on local neighbourhood node features.
    \item[--] Attention link function: Attention-based link function to predict the adjacent matrix of a graph.
    \item[--] Surgical scene graph dataset: Annotate bounding box and tool-tissue interaction to predict scene graph during robotic surgery.
\end{itemize}

\section{Proposed Approach}
\begin{figure}[!h]
    \centering
    \includegraphics[width=1\textwidth]{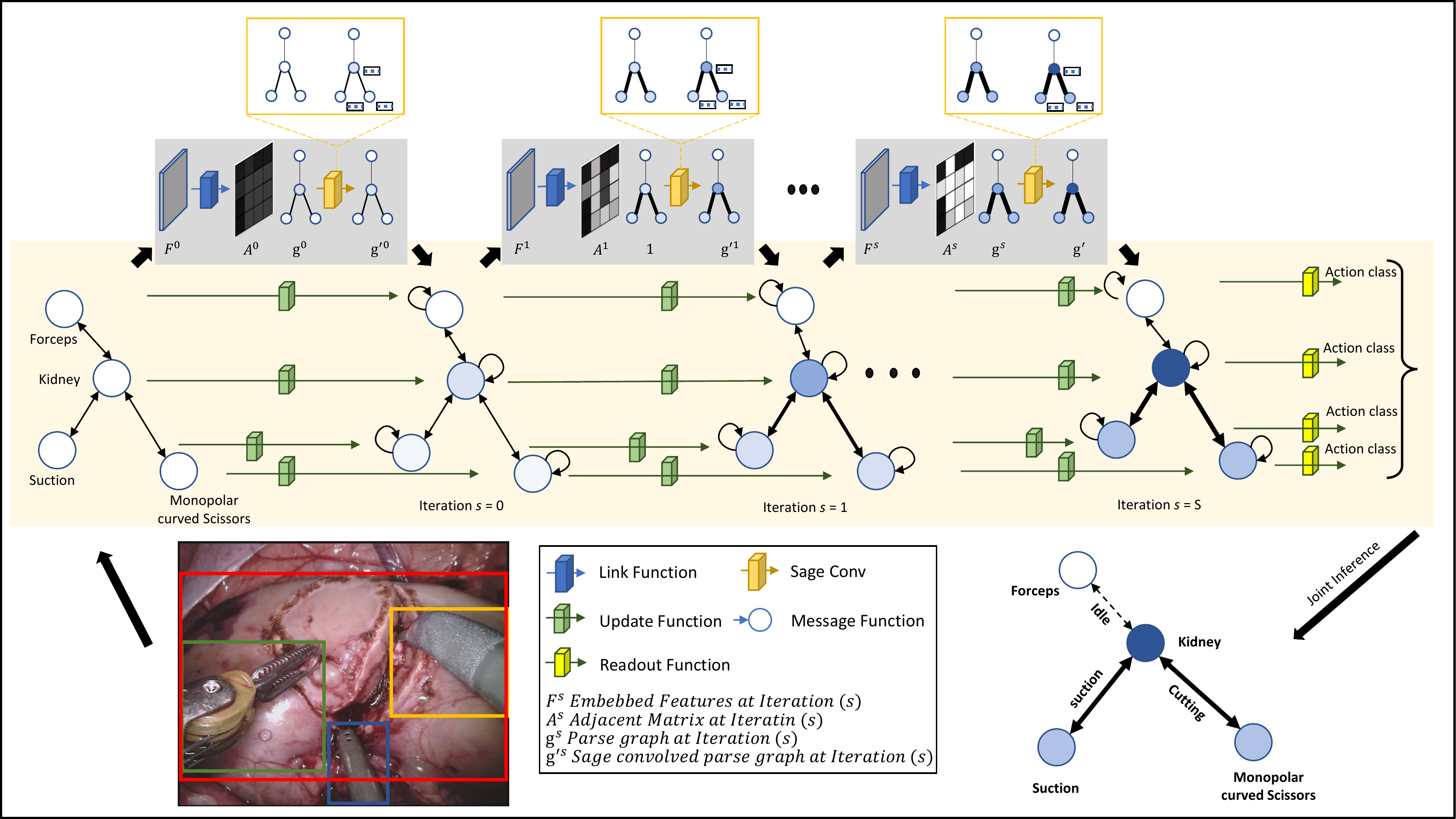}
    \caption{Proposed architecture: Given a surgical scene, firstly, label smoothened features $\mathcal{F}$ are extracted. The network then outputs a parse graph based on the $\mathcal{F}$. The attention link function predicts the adjacent matrix of the parse graph. The thicker edge indicates possible interaction between the node. The SageConv embeds each node with its neighbouring node features. A brighter node color represents the existence of neighbouring node features within a node. The message function updates the node values based on the sum of interaction features between neighbour nodes. The hidden features of each node are updated by the update node. The readout function predicts the node labels that signify the interaction type between the connected nodes.}%
    \label{fig:architecture}
\end{figure}

To perform surgical scene understanding, the interactions between the defective tissue and surgical instruments are presented in the non-euclidean domain. Surgical tools, defective tissues and their interactions are represented as node and edge of a graph structure using the label smoothened spatial features extracted using the ResNet18\cite{he2016deep} (as Section \ref{LS_Feature}). Inspired from GPNN\cite{qi2018learning}, the complete tissue-tool interaction is modelled in the form of graph $\mathcal{G}(\mathcal{V},\mathcal{E},\mathcal{Y})$. The nodes $v \in \mathcal{V}$ of the graph signifies the defective tissue and instruments. The graph's edges $e \in \mathcal{E}$ denoted by $e=(x,w) \in \mathcal{V}$ x $\mathcal{V}$ indicates the presence of interaction. The notation $y_v \in \mathcal{Y}$ correlates to a set of labels, holding the interaction state of each node $v \in \mathcal{V}$.

Each image frame in the surgical scene is inferred as a parse graph $g = (\mathcal{V}_g,\mathcal{E}_g,\mathcal{Y}_g$), where, $g$ is a subgraph of $\mathcal{G}, \mathcal{E}_g \subseteq \mathcal{E}$ and $\mathcal{V}_g \subseteq \mathcal{V}$ as shown in Fig.\ref{fig:architecture}. Given a surgical scene, the SSD network is first employed to detect bounding boxes. Label smoothened features $\mathcal{F}$, comprising of node features $\mathcal{F}_v$  and edge features $\mathcal{F}_e$ are then extracted based on these detected bounding boxes. Utilizing these features, an enhanced graph-based deep reasoning model then infers a parse graph, $\displaystyle g^* $ = $argmax_{g}$ $p(\mathcal{Y}_g|\mathcal{V}_g,\mathcal{E}_g, \mathcal{F})$ $p(\mathcal{V}_g,\mathcal{E}_g|\mathcal{F},\mathcal{G})$ \cite{qi2018learning} to deduce interactions between the defective tissue and surgical tools. The default GPNN model features four functions. The link function that consists of a convolution and relu activation layer predicts the structure of the graph. The message functions summaries the edge features for each node using fully connected layers. The update function features a gated recurrent unit (GRU) that utilizes the summarised features to update the hidden node state.  At the end of the propagation iteration, the readout functions use fully connected layers to predict the probability of interaction for each node. Here, the model additionally incorporates SageConv and spatial attention module \cite{roy2018concurrent} in the link function. The sage convolution embeds each node with its neighbouring node features. The attention module helps increase module accuracy by amplifying significant features and suppressing weak features.

\subsection{Label Smoothened Feature}
\label{LS_Feature}

\begin{figure}[!htbp]
    \centering
    \includegraphics[width=1\textwidth]{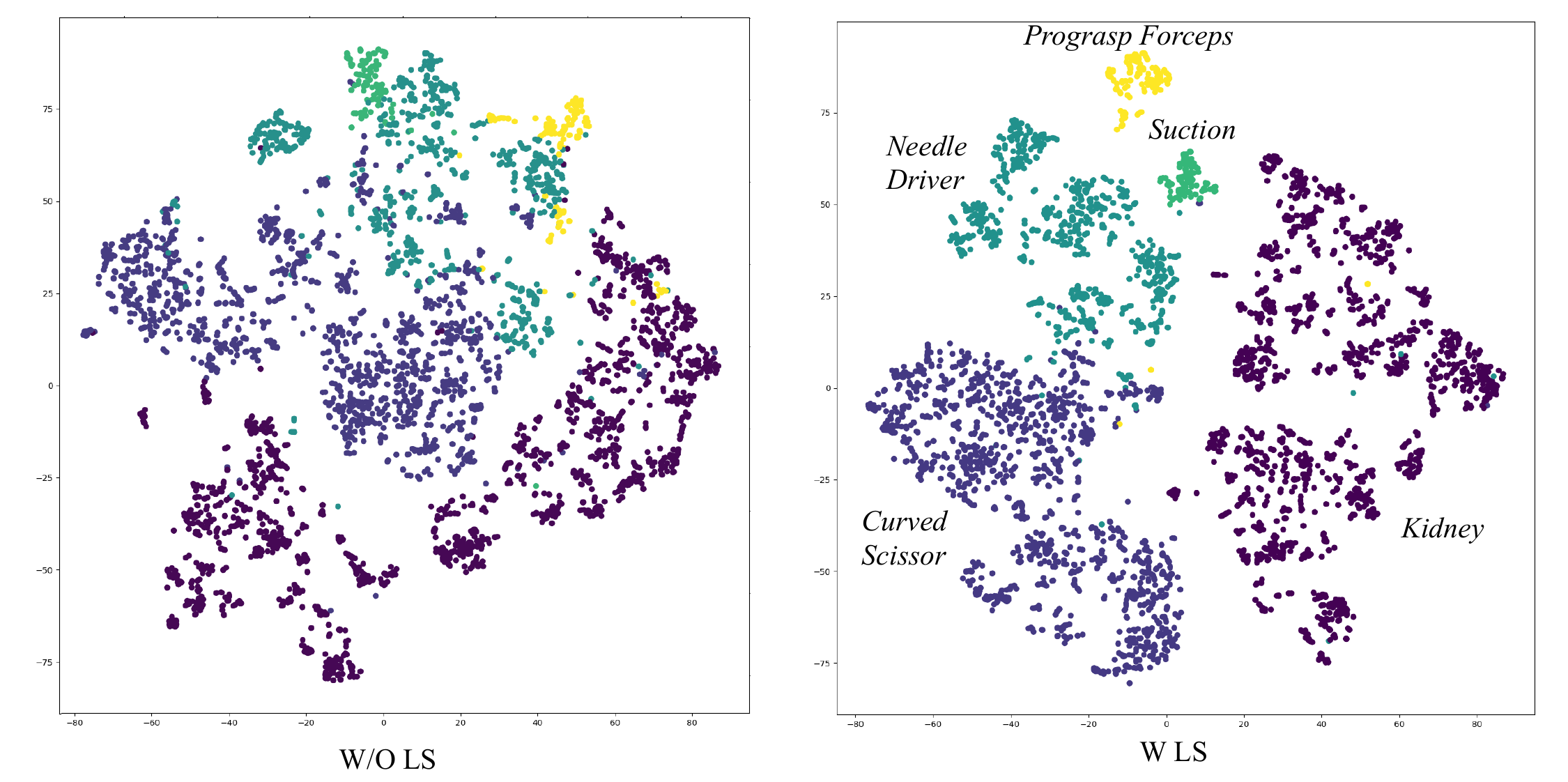}
    \caption{Node feature extraction with and without label smoothing (LS). We choose 5 semantically different classes to plot the tSNE. Based on the comparison between the two images, it is observed that the clusters containing features obtained from the same class are more compact when extracted using the LS method.}%
    \label{fig:graph_embedding}
\end{figure}

Label smoothing (LS) calculates the soft-target by weighting the true target and use it to measure the cross-entropy loss. For example, if $T_k$ is the true onehot target and $\epsilon$ is the smoothing factor then smoothen label for the K classes, $T_k^{LS} = T_k(1-\epsilon) + \epsilon/K$ and LS cross-entropy loss for the model prediction $P_k$, $CE^{LS} = \sum_{k=1}^{K} -T_k^{LS} log (P_k)$. It is observed that LS prevents the over-confident of model learning and calibrates the penultimate layer and represents the features of the same classes in a tight cluster \cite{muller2019does}. We employ LS on ResNet18 to train with surgical object classification and extract edge and node features for graph learning. With LS, the cluster of the same classes is much tighter than the extracted features without LS. The label smoothened features are represented by $\mathcal{F}$ = {$\mathcal{F}_v$, $\mathcal{F}_e$}, where $\mathcal{F}_v$ and $\mathcal{F}_e$ denotes the label smoothened node and edge features respectively.


\subsection{Inductive graph learning by SageConv}
SageConv\cite{hamilton2017inductive}, an inductive framework, enables node embedding by sampling and aggregating from the node's local neighbourhood. It is incorporated into the model to introduce inductive graph learning in the parse graph. Based on the adjacent matrix deduced by the link function, the SageConv further embeds each node in the parse graph by sampling and aggregating its connected neighbouring node features. Given an independent node feature $\mathcal{F}_v$, $\forall v \in \mathcal{V}$, the SageConv outputs a new embedded node features $\mathcal{F}_v \leftarrow \mathcal{F}^{K}_v$, $\forall v \in \mathcal{V}$ where $\mathcal{F}^{K}_v$ denotes the aggregated node features from the connected neighbourhood($K$) nodes.

\section{Experiments}

\subsection{Dataset}

\begin{figure}[!h]
    \centering
    \includegraphics[width=1\textwidth]{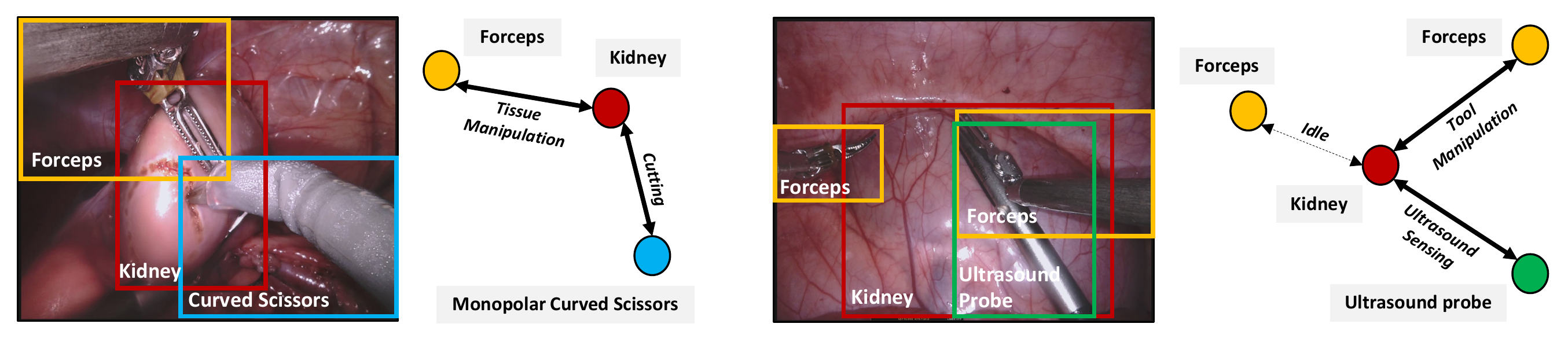}
    \caption{Graph annotation for tool-tissue interaction in a robotic surgical scene: Each surgical scene is annotated using a graph structure. Each node is the graph denotes a defective tissue or surgical instrument. The edges between the nodes represent the type of interaction between the nodes.}
    \label{scene_graph_annotation}
\end{figure}

In this work, the robotic scene segmentation challenge 2018 dataset \cite{allan20202018} is exploited to generate a graph-based tissue-tool interaction dataset. The training subset consists of 15 robotic nephrectomy procedures captured on the da Vinci X or Xi system. There are 149 frames per video sequence, and the dimension of each frame is 1280x1024. Segmentation annotations are provided with 10 different classes, including instruments, kidneys, and other objects in the surgical scenario. The main differences with the 2017 instrument segmentation dataset \cite{allan20192017} are annotation of kidney parenchyma, surgical objects such as suturing needles, Suturing thread clips, and additional instruments. We annotated the graphical representation of the interaction between the surgical instruments and the defective tissue in the surgical scene with the help of our clinical expertise with the da Vinci Xi robotic system. We also delineate the bounding box to identify all the surgical objects. Kidney and instruments are represented as nodes and active edges annotated as the interaction class in the graph. In total, 12 kinds of interactions were identified to generate the scene graph representation. The identified interactions are grasping, retraction, tissue manipulation, tool manipulation, cutting, cauterization, suction, looping, suturing, clipping, staple, ultrasound sensing. We split the newly annotated dataset into 12/3 video sequences (1788/449 frames) for training and cross-validation.


\subsection{Implementation Details}
We choose sequences $1{st},5{th}$ and $16{th}$ for the validation and remaining 12 sequences for the training. To train our proposed model, weighted multi-class multi-label hinge loss \cite{moore2011l1, qi2018learning}, and adam optimizer with a learning rate of \(10^{-5}\) are employed. Pytorch framework with NVIDIA RTX 2080 Ti GPU is used to implement all the codes.
\subsubsection{Graph Feature Extraction}
To extract node and edge features from images, there are extensive experiments we have conducted in this work. We select 4 different types of deep neural networks such as classification, detection, segmentation and multitask learning models and train them with corresponding annotations for both w and w/o LS in cross-entropy loss. As instrument segmentation challenge 2017 \cite{allan20192017} consists of similar instruments and surgical activities, we choose pre-trained models to initialize the segmentation and detection network from the previous works \cite{islam2020ap, islam2019learning}. Further, the trained model is utilized to extract node and edge features by adding adaptive average pooling after the penultimate layer as \cite{muller2019does, qi2018learning}. We choose feature vector of size 200 for each node and edge after tuning. The extracted features are employed to train the proposed graph parsing network for scene generation and interaction prediction.

\subsection{Results}

\begin{figure}[!h]
    \centering
    \includegraphics[width=1\textwidth]{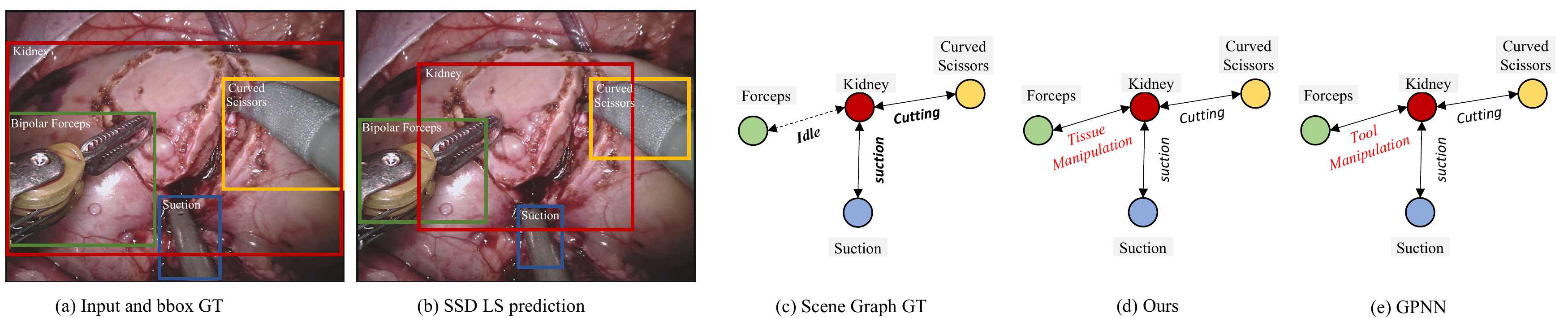}
    \caption{Qualitative analysis: (a) Input surgical scene with true bounding box. (b) Bounding box prediction based on SSD LS model. (c) Scene graph ground truth that represents the tissue-tool interaction in the scene. (d) Predicted scene graph based on our proposed model. (e) Predicted scene graph based on GPNN model \cite{qi2018learning}. The interaction highlighted in red in (d) and (e) indicates the false prediction made by the respective models.}
    \label{fig:qualitative_comparison}
\end{figure}

The enhanced network proposed in the model is evaluated both qualitatively and quantitatively against the state of the art networks. Fig \ref{fig:qualitative_comparison} shows the qualitative performance of the SSD LS model (b) in predicting the bounding box against the ground truth (a). The figure also shows our model's (d) scene graph prediction against the ground truth (GT) (c) and the base GPNN model (e). Although both models made a wrong interaction prediction between the kidney and the forceps, based on the spatial features observed, it is arguable that our model's prediction was more reasonable compared to the GPNN model. Qualitative comparison with GraphSage \cite{hamilton2017inductive}, GAT \cite{velivckovic2017graph} and Hpooling \cite{zhang2019hierarchical} models were not performed as these models are unable to predict the adjacent matrix (presence of interaction) between the graph nodes.  

\begin{table}[!h]
\begin{center}
\caption{Comparison of our proposed graph model's performance against the state of the art model in scene graph generation. RO. denotes ReadOut module.}
\label{table:quantitative_comparison}
\begin{tabular}{l|c||c|c|c|c|c} 
\bottomrule[1.4px]
\multicolumn{2}{c||}{Detection}    &  \multicolumn{5}{c}{Graph Scene Prediction @ 0.5479mAP}  \\ \hline

Models & mAP$\uparrow$ & Models      & mAP $\uparrow$ & Hinge $\downarrow$   & Recall $\uparrow$ & AUC $\uparrow$ \\ \hline
SSD \cite{liu2016ssd} w LS &\textbf{0.5479} & \textbf{Ours}        & \textbf{0.3732}       &\textbf{1.74}   & 0.7989       &0.8789      \\ \hline
SSD \cite{liu2016ssd} w/o LS &0.5290 & GPNN \cite{qi2018learning} &0.3371  &1.79   &0.8001  &0.8782       \\ \hline
YoloV3\cite{redmon2018yolov3} w LS&0.4836 & GraphSage \cite{hamilton2017inductive} + RO.  &0.2274          &3.10                &0.6957             & 0.7468      \\ \hline
YoloV3\cite{redmon2018yolov3} w/o LS&0.4811 & GAT \cite{velivckovic2017graph}        &0.1222          &16.14               &0.7577             &0.7866 \\ \hline
-&- & Graph HPooling \cite{zhang2019hierarchical}   &0.1883          &13.93                     &0.7727             &0.8282  \\ 
\toprule[1.4px]
\end{tabular}
\end{center}
\end{table}

Quantitative comparison results between the proposed model and the state of the art model are reported in table \ref{table:quantitative_comparison}. Firstly, the advantage of incorporating the label smoothing technique is reported under the detection. When compared with the default YoloV3\cite{redmon2018yolov3} and SSD\cite{liu2016ssd} detection models against incorporating them with the label smoothing technique, an increase in mean average precision (mAP) is reported. Secondly, the performance of our overall proposed model against graph-based models is also shown in Table \ref{table:quantitative_comparison}. When compared with the default GPNN \cite{qi2018learning}, GraphSage \cite{hamilton2017inductive} (incorporated with Readout function from \cite{qi2018learning}), GAT \cite{velivckovic2017graph} and Graph Hpooling, our enhanced model achieves a higher mAP and marginally better area under the curve. However, the GPNN \cite{qi2018learning} has a marginally better performance in terms of hinge loss and recall.


\begin{table}[!h]
\centering
\caption{Ablation study of proposed model while integrating different methods and modules.}
\label{table:module_ab}
\begin{tabular}{c|c|c|c|c|c|c}
\toprule[1.4px]
\multicolumn{4}{c|}{\textbf{Modules and Methods}} &\multirow{1}{1cm}{\textbf{mAP}} & \multirow{1}{1cm}{\textbf{Hinge}} & \multirow{1}{1cm}{\textbf{AUC}} \\ \cline{1-4}
\textbf{Base} & \textbf{LS} &\textbf{Attention} & \textbf{SageConv} &  & & \multicolumn{1}{l}{} \\ \hline
\cmark & \cmark & \cmark & \cmark &0.3732  &1.74  &0.8789 \\ \hline
\cmark & \cmark & \cmark & \xmark &0.3662   &1.70  &0.8699 \\ \hline
\cmark & \cmark & \xmark & \xmark &0.3371   &1.79  &0.8782  \\ \hline
\cmark & \xmark & \xmark & \xmark &0.3145  &3.31  &0.8122   \\ \bottomrule[1.4px]

\end{tabular}
\end{table}

\begin{figure}[!htbp]
    \centering
    \includegraphics[width=0.7\textwidth]{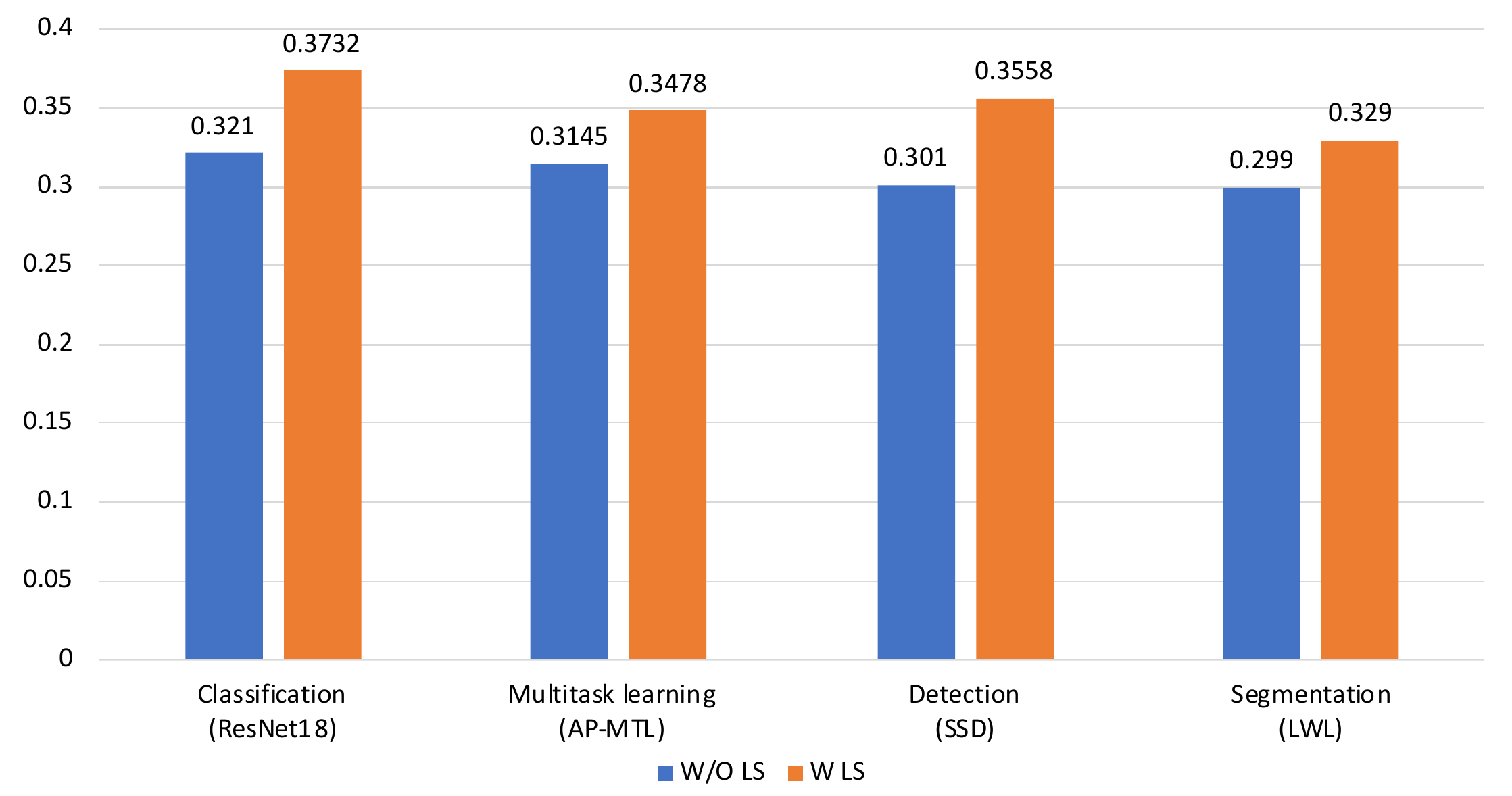}
    \caption{Mean average precision (mAP) of the scene graph prediction with various feature embedding models such as classification, detection and segmentation. Our experiments confirm that classication model with LS produces highest accuracy. We exploit ResNet18 \cite{he2016deep}, AP-MTL \cite{islam2020ap}, SSD \cite{liu2016ssd} and LWL \cite{islam2019learning} models to extract the feature from the penultimate layer.}
    \label{fig:ls_feature}
\end{figure}

The ablation study of the proposed model is shown in table \ref{table:module_ab}. The table relates the increase in module's performance to every module added onto the default GPNN base module. The use of label smoothened features has significantly increased the base model's performance in terms of mAP, hinge loss and AUC. Additional attention link function has further increased the accuracy of the model. Further additional of SageConv has increased both the mAP and AUC performance of the model. Fig.\ref{fig:ls_feature} demonstrates our models performance on various type of deep neural network model such as classification (ResNet18), multitask learning (AP-MTL \cite{islam2020ap}), detection (SSD \cite{liu2016ssd}) and segmentation (LWL \cite{islam2019learning}). It is interesting to see that the classification model extracts the most distinguishable features comparing to other methods. The figure also infers that extracted features with LS have better representation than without LS.

\section{Discussion and Conclusion}
In this work, we propose an enhanced graph-based network to perform spatial deep reasoning of the surgical scene to predict the interaction between the defective tissue and the instruments with an inference speed of 15 ms. We design and attention link function and integrated SageConv with a graph parsing network to generate a scene graph and predict surgical interactions during robot-assisted surgery. Furthermore, we demonstrate the advantage of label smoothing in graph feature extraction and the enhancement of the model performance. Spatio-temporal scene graph generation and relative depth estimation from sequential video in the surgical environment can be the future direction of robotic surgical scene understanding.

\bibliography{mybib}{}
\bibliographystyle{splncs03}

\clearpage
\end{document}